# SCUT-FBP: A Benchmark Dataset for Facial Beauty Perception

Duorui Xie, Lingyu Liang, Lianwen Jin*, Jie Xu, Mengru Li
School of Electronic and Information Engineering
South China University of Technology, Guangzhou 510640, China
*Email: lianwen.jin@gmail.com

*Abstract*—In this paper, a novel face dataset with attractiveness ratings, namely, the SCUT-FBP dataset, is developed for automatic facial beauty perception. This dataset provides a benchmark to evaluate the performance of different methods for facial attractiveness prediction, including the state-of-the-art deep learning method. The SCUT-FBP dataset contains face portraits of 500 Asian female subjects with attractiveness ratings, all of which have been verified in terms of rating distribution, standard deviation, consistency, and self-consistency. Benchmark evaluations for facial attractiveness prediction were performed with different combinations of facial geometrical features and texture features using classical statistical learning methods and the deep learning method. The best Pearson correlation (0.8187) was achieved by the CNN model. Thus, the results of our experiments indicate that the SCUT-FBP dataset provides a reliable benchmark for facial beauty perception.

## I. Introduction

Assessing facial beauty is a challenging task that has been investigated by countless philosophers, artists, and scientists for many years. In particular, it has attracted considerable attention in the field of computer vision. Recent psychology research [1] has shown that the perception of beauty is consistent among different individuals. Another study [2] has indicated that facial beauty is a universal concept that can be learned by a machine. Research on facial beauty, which can serve as the basis for facial aesthetics, plastic surgery, and face image retouching, has contributed to the development of commercial systems for facial beauty enhancement, such as MeiTu [24] and Portraiture [25].

Most studies on facial beauty focus on designing facial beauty descriptors. Because facial symmetry, averageness, and secondary sex characteristics influence the perception of facial attractiveness [5, 6], data-driven facial beauty analysis based on geometric features [3, 4, 9] and skin texture features [7] has inspired many related studies in the fields of computer vision and machine learning. Although feature extraction for facial beauty analysis has been investigated extensively, little attention has been paid to data collection in this regard. A publicly available facial beauty dataset is expected to facilitate further research in this field. In particular, it can provide a unified benchmark for evaluating the performance of different algorithms, thereby promoting the development of new algorithms and applications for facial beauty analysis as well as selection criteria for facial beautification [32].

Many studies on facial attractiveness prediction [8, 19] have used existing face databases for evaluation, such as the databases for face recognition and smile detection [29]. Although these databases are suitable for some specific face analysis task, they may fail to meet the requirements of the facial beauty perception problem owing to the lack of attractiveness ratings.

Face datasets [12-13, 17] for facial beauty assessment were built in a recent study. Fan et al. proposed a dataset [12] containing computer-generated face images with different facial proportions; however, its use is limited for face structure analysis. Yan [13] proposed dataset gathering from social networks, but the resolution of the collected images was low. There are some large-scale databases for facial beauty analysis, such as the Northeast China database [4], the Shanghai database [9], and the recent AVA database [15], which can be improved in certain aspects from the perspective of facial beauty perception. The Northeast China database [4] and the Shanghai database [9] are limited for geometric facial beauty analysis; they fail to capture the appearance features and the corresponding attractiveness ratings. The AVA database [15], a large-scale database for aesthetic visual analysis, contains a subset of portraits [14]. However, AVA is concerned with the aesthetic analysis of the entire image and not just the face. Therefore, the AVA ratings of a portrait reflect the quality of the image but not of the face itself; thus, a portrait may have a high rating because of the background or facial expressions, and not because of the attractiveness of the face.

TABLE I. SOME REPRESENTATIVE DATASETS FOR FACIAL BEAUTY ANALYSIS

| Dataset | Number of Images | Raters per Image | Beauty Class | Publicly Available or Not |
|---|---|---|---|---|
| [2] | 92/92 | 28/18 | 7 | NO |
| [4] | 23412 | unknown | 2 | NO |
| [9] | 1307 | 100 | unknown | NO |
| [17] | 215 | 46 | 10 | NO |
| [12] | 432 | 30 | 7 | NO |
| [14] | 10141 | 78–549 | 10 | YES |
| SCUT-FBP | 500 | 70 | 5 | YES |

This paper proposes a benchmark dataset, namely, the SCUT-FBP dataset, which can be applied to different facial beauty analysis problems, including facial attractiveness and

facial beautification. The main contributions of this paper can be summarized as follows:

- **Dataset.** A large number of portraits with different levels of attractiveness are collected. To reduce the effects of irrelevant factors, SCUT-FBP contains high-resolution, front-on face portraits of Asian female subjects with neutral expressions, simple backgrounds, and minimal occlusion; these factors are conducive to facial beauty perception in both geometry and appearance.

- **Beauty Rating Analysis.** Attractiveness ratings for all images are collected, and the final rating is determined according to the rating distribution. The average number of raters per image of the SCUT-FBP dataset is 70, which is greater than that of the datasets used in previous studies [9, 11, 12, 17]. We verify the ratings in terms of the rating distribution [14], standard deviation [14], consistency [2], and self-consistency [19].

- **Feature Analysis.** We propose the use of an 18-dimensional geometrical feature and 2-dimensional Gabor texture features to predict facial attractiveness. The 18-dimensinal geometrical feature is based on traditional Chinese facial beauty standards. To extract texture features, we adopt two sampling methods that reduce the dimension and enhance the accuracy of the prediction. Experiments show that the above-mentioned features can represent facial beauty with sufficient accuracy.

- **Beauty Prediction.** Both traditional machine-learning and deep learning methods are adopted to predict beauty. The best Pearson correlation for traditional machine learning and deep learning is 0.6482 and 0.8187, respectively, which indicates that the SCUT-FBP dataset provides a reliable benchmark for facial beauty analysis.

The remainder of the paper is organized as follows. Section II describes the creation of the SCUT-FBP dataset. Section III discusses the analysis of the dataset. Section IV presents benchmark evaluations of the dataset. Finally, Section V concludes the paper by summarizing our findings.

## II. CREATION OF SCUT-FBP

### A. Data collection

We collected data to build a standard dataset that provides unified data for evaluating the performance of different algorithms. To reduce the effects of irrelevant factors such as age, gender, and facial expression, the SCUT-FBP dataset is confined to a unified form, i.e., it contains high-resolution, front-on face portraits of Asian female subjects with neutral expressions, simple backgrounds, no accessories, and minimal occlusion. A previous study [20] has shown that beautiful individuals constitute a small percentage of the population. The SCUT-FBP dataset contains a higher proportion of beautiful faces than that in the general population in order to facilitate effective learning of facial beauty. Specifically, it contains 500 portraits, some of which we captured ourselves; others were licensed from different sources [26-28] or downloaded from the Internet. All the images were rated by numerous raters. Figure 1 shows some examples of face portraits from the dataset.

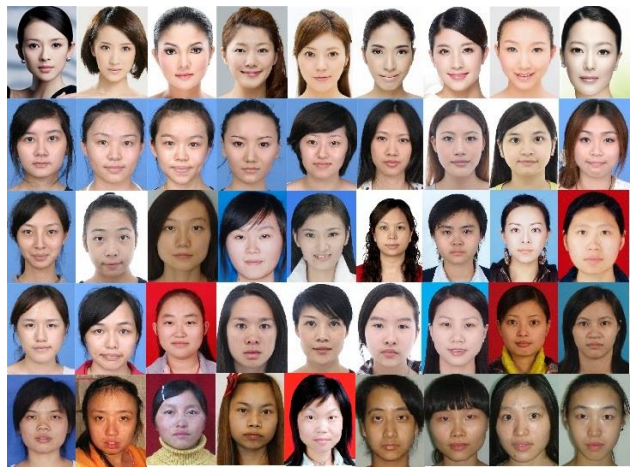

Figure 1. Examples with different levels of beauty in the SCUT-FBP dataset, which is publicly available at http://www.hcii-lab.net/data/SCUT-FBP/

### B. Rating collection

We developed a web-based tool, namely, the facial beauty assessment system, to collect ratings. Images in the SCUT-FBP dataset were rated by 75 raters; the average number of raters per image was 70. Because the evaluation ground truth varied among individuals, we obtained raters' opinions regarding the beauty of the portraits by asking them for answers to certain questions [10, 31]. The questions are listed in Figure 2. The portraits were randomly shown to the raters. The raters could change their ratings if they accidentally selected an incorrect option. Although facial beauty has been shown to be a universal concept [2], it is subjective to some extent. The procedure described above aims to eliminate unnecessary effects.

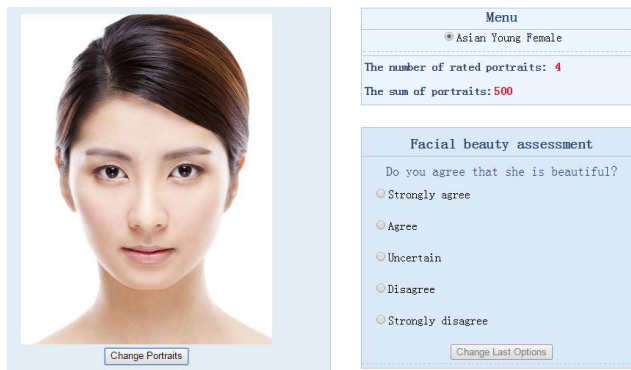

Figure 2. Interface of our facial beauty assessment system[1]

---

[1] The facial beauty assessment system can be accessed online at http://202.38.194.248:8011/.

The rating process is summarized as follows:

- 75 raters were invited to use the facial beauty assessment system and rate the portraits.
- The system displayed the portraits in a random manner.
- The raters could rate a portrait or change the rating given to the last viewed portrait by clicking the "Change Last Operation" button. In addition, they could view the next portrait by clicking the "Change portrait" button (see Figure 2).
- We analyzed the ratings, selected the appropriate data, and omitted the erroneous data. Then, we plotted a histogram for every portrait. The average rating of all the raters was defined as the attractiveness rating label.

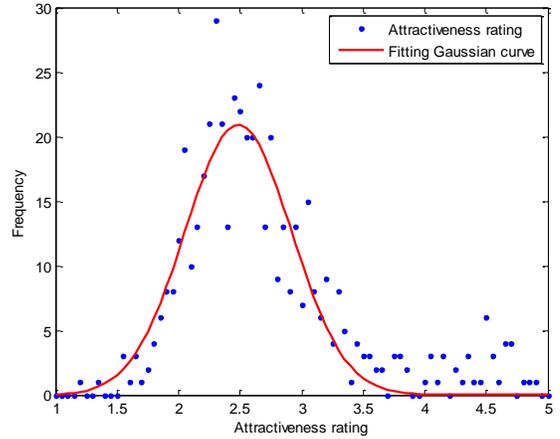

Figure 4. Histogram of rating distribution

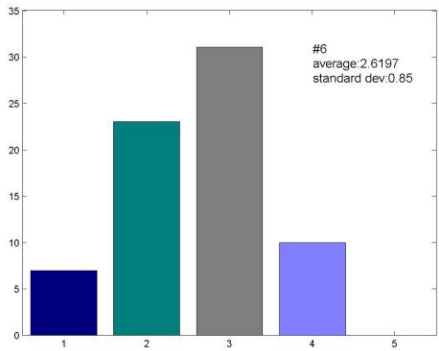

Figure 3. Example of grade histogram for image 6 from the dataset

- To verify the self-consistency of the raters, we invited 20 raters (10 male, 10 female) from among the original raters for re-rating after 1 week and 2 weeks; these rating sets were used for verifying self-consistency.

### III. ANALYSIS OF SCUT-FBP

In this section, we describe the analysis of the SCUT-FBP dataset in terms of the following aspects: rating distribution, standard deviation, consistency, and self-consistency.

#### A. Rating distribution

We statistically analyzed the rating distribution for the entire dataset. The histogram of the rating distribution is shown in Figure 4.

This figure shows that the rating distribution is nearly Gaussian. The major part of the dataset consists of portraits having an average rating of around 2.5. This implies that average faces are more common than beautiful and unattractive faces, which reflects the real-world situation. In Figure 4, there is a small peak around 4.5 because the dataset contains a higher proportion of beautiful faces than the general population in order to facilitate effective learning of facial beauty. The rating distribution is consistent with our expectation.

#### B. Standard deviation

The standard deviation of the ratings indicates the raters' consistency: a low standard deviation denotes high consistency. The standard deviation is concentrated between 0.6 and 0.8. The highest standard deviation is 1.07, the lowest standard deviation is 0.41, and the average standard deviation is 0.693. A small standard deviation indicates high consistency in the perception of facial beauty, thus verifying the rationality of our rating label set.

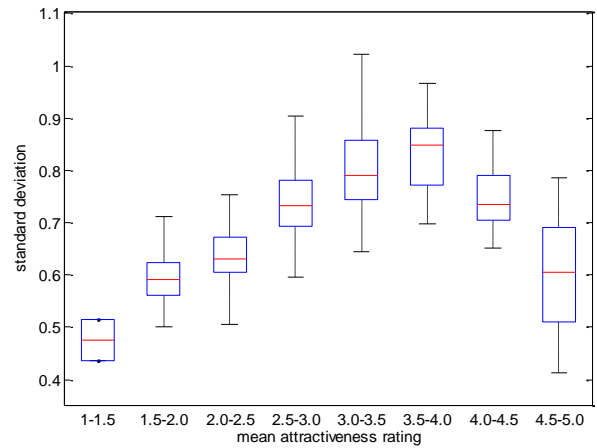

Figure 5. Distribution of standard deviation for portraits with different mean ratings

Figure 5 shows the plot of standard deviation for portraits with mean ratings within a specific range. From this figure, it can be seen that portraits with average ratings (ratings in the range (2.5, 3.5)) tend to have a higher standard deviation than portraits with ratings greater than 3.5 or less than 2.5. The closer the score to 1 or 5, the lower is the standard deviation. There same conclusion is reached in the case of AVA [15] This indicates that there is a unified opinion regarding a beautiful face and an unattractive face, but the perception of an average face is rather subjective.

## C. Consistency

Previous studies [2, 17, 11] divided ratings into two groups, calculated the mean rating of each group, and checked for consistency between the two mean ratings. We repeated this procedure numerous times. The correlation between the two mean ratings was found to be 0.966–0.973, which was higher than the correlations obtained previously (0.9–0.95 [2] and 0.87–0.9 [11]).

The t-test has also been used for dataset verification [2, 17]. We employed the t-test in our experiment and found that the mean ratings of the two groups were not statistically different.

## D. Self-consistency

Three sets of ratings were collected over different periods. We check these sets for consistency.

Table II lists the self-consistency correlations for 20 individuals. The average correlation was 0.65–0.85. Further, the self-consistency of females (0.739) was slightly higher than that of males (0.714). The average correlation for 20 raters was 0.727, which was higher than that obtained previously (0.58 [19]).

For the entire dataset, the self-consistency correlations among the three sets were 0.9704, 0.9705, and 0.9758, which represents a strong correlation.

In general, the self-consistency of both the raters and the entire dataset was high, which confirms the reliability of our rating data.

TABLE II. CORRELATIONS BETWEEN DIFFERENT SETS OF LABELS OF ATTRACTIVENESS

| Rater # | $1^{st}$-$2^{nd}$ correlation | $2^{nd}$-$3^{rd}$ correlation | $1^{st}$-$3^{rd}$ correlation | Average correlation |
|---|---|---|---|---|
| 1(female) | 0.68 | 0.74 | 0.69 | 0.70 |
| 2(female) | 0.64 | 0.68 | 0.69 | 0.67 |
| 3(female) | 0.70 | 0.68 | 0.61 | 0.67 |
| 4(female) | 0.73 | 0.73 | 0.71 | 0.72 |
| 5(female) | 0.71 | 0.73 | 0.87 | 0.77 |
| 6(female) | 0.73 | 0.68 | 0.72 | 0.71 |
| 7(female) | 0.78 | 0.80 | 0.81 | 0.80 |
| 8(female) | 0.79 | 0.80 | 0.83 | 0.81 |
| 9(female) | 0.85 | 0.86 | 0.85 | 0.85 |
| 10(female) | 0.66 | 0.67 | 0.72 | 0.69 |
| Average correlation for females | | | | 0.739 |
| 11(male) | 0.77 | 0.76 | 0.78 | 0.77 |
| 12(male) | 0.73 | 0.69 | 0.73 | 0.72 |
| 13(male) | 0.72 | 0.73 | 0.75 | 0.73 |
| 14(male) | 0.69 | 0.66 | 0.69 | 0.68 |
| 15(male) | 0.71 | 0.71 | 0.85 | 0.76 |
| 16(male) | 0.68 | 0.72 | 0.76 | 0.72 |
| 17(male) | 0.71 | 0.67 | 0.64 | 0.67 |
| 18(male) | 0.67 | 0.66 | 0.61 | 0.65 |
| 19(male) | 0.63 | 0.67 | 0.67 | 0.66 |
| 20(male) | 0.76 | 0.78 | 0.81 | 0.78 |
| Average correlation for males | | | | 0.714 |
| Average for 20 raters | | | | 0.727 |

## IV. BENCHMARK EVALUATION

Facial attractiveness rating can be regarded as a regression problem. This section describes benchmark evaluations conducted by comparing traditional machine learning and deep learning.

### A. Facial beauty prediction using traditional machine learning methods

Using traditional machine learning, we aimed to develop a suitable feature extraction and machine-learning algorithm in order to learn and predict beauty automatically.

*1) Feature extraction:* We use the geometric features and skin texture features that have been employed in several previous studies [4, 9, 22].

*a) Geometric features:* As shown in Figure 6, we extracted 18 features to abstractly represent each face based on [3]. In addition to the 17 features in [3], we included the vertical distance from the hairline to the midpoint between the eyebrows.

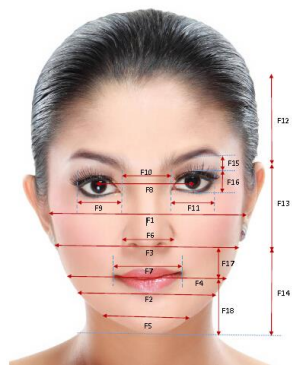

Figure 6. Example showing 18 geometric features

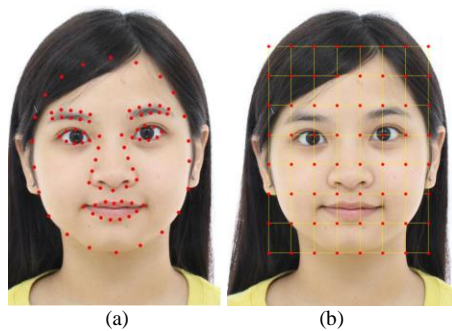

(a)      (b)

Figure 7. Two different sampling methods

*b) Skin texture features:* A study [7] has shown that skin texture plays a significant role in the perception of female facial beauty. A Gabor filter with 4 scales and 8 directions was applied to extract texture. Two sampling methods (see Figure 7) were adopted to extract skin texture information. The first sampling method is shown in Figure 7(a). We extracted 84 points as sample points containing facial contour information and shape information of the eyebrow, eyes, mouth, and so on. The second sampling method is shown in Figure 7(b). We selected the smallest rectangle that can include a face region. Then, 8 × 8 uniform sampling was conducted within this rectangle. The 64 points were collected as sample points. The Gabor features around the sample points, KeyPointGabor and UniSampleGabor, represent the face.

### 2) Facial beauty prediction

*a) Performance based on geometric features:* In this subsection, we evaluate the prediction performance of different algorithms on the basis of several criteria such as Pearson correlation (PC) [30], mean absolute error (MAE) [10], and root mean squared error (RMSE) [10]. The machine-learning methods we used include SVM regression (SVR), linear regression, pace regression, and Gaussian regression.

TABLE III. PREDICTION PERFORMANCE OF DIFFERENT ALGORITHMS BASED ON GEOMETRIC FEATURES

| Regression algorithm | *Linear regression* | *Pace regression* | *Gaussian regression* | *SVR* |
|---|---|---|---|---|
| PC | 0.5921 | 0.5847 | 0.6057 | 0.608 |
| MAE | 0.412 | 0.4139 | 0.4014 | 0.4021 |
| RMSE | 0.5389 | 0.5422 | 0.5316 | 0.5316 |

From Table III, it can be seen that the best Pearson correlation (0.608) was achieved by SVR. Gaussian regression also showed good performance. Therefore, in the following experiments, we adopted Gaussian regression and SVR algorithms.

*b) Performance based on texture features:* Principal component analysis (PCA) was adopted to reduce the high dimension of the extracted Gabor features.

From Table IV, we can see that the skin texture feature sampled in the second method showed better performance than that in the first method (Pearson correlation of 0.6347 based on Gaussian regression).

TABLE IV. PREDICTION PERFORMANCE OF DIFFERENT ALGORITHMS BASED ON TEXTURE FEATURES

| | KeyPointGabor + PCA | | UniSampleGabor + PCA | |
|---|---|---|---|---|
| | *SVR* | *Gaussian regression* | *SVR* | *Gaussian regression* |
| PC | 0.549 | 0.4591. | 0.5847 | 0.6347 |
| MAE | 0.5541 | 0.4724 | 0.423 | 0.3969 |
| RMSE | 0.5606 | 0.6152 | 0.5452 | 0.5164 |

*c) Performance based on combination of texture and geometric features:* We combined the geometric and UniSampleGabor features, referred to as the combined feature, in order to improve prediction performance. Gaussian regression showed the best performance (Pearson correlation, 0.6482). The combined feature showed better performance than the individual features, which indicates that both geometric features and skin texture are important for the perception of facial beauty.

TABLE V. PREDICTION PERFORMANCE OF DIFFERENT ALGORITHMS BASED ON COMBINED FEATURE

| Algorithm | PC | MAE | RMSE |
|---|---|---|---|
| SVR | 0.6433 | 0.3961 | 0.512 |
| Gaussian regression | 00.6482 | 0.3931 | 0.5149 |

### B. Facial beauty prediction using deep learning

Deep learning is a new area of machine learning. It sets up a network that can mimic the human brain for thinking and learning tasks. A traditional approach to facial beauty prediction involves extracting features from images manually and adding them into a classifier for classification. Such an approach is inefficient and highly dependent on operator experience. In contrast, deep learning combines feature extraction and classification so that features can be learned automatically from the input data.

Deep learning attempts to learn in multiple levels corresponding to different levels of abstraction. The levels in these learned statistical models correspond to distinct levels of concepts, where higher-level concepts are defined from lower-level concepts, and the same lower-level concepts can be used to define many higher-level concepts.

A convolutional neural network (CNN) is an important framework of deep learning. It consists of various combinations of convolutional layers, pooling layers, and fully connected layers. Such a structure allows a CNN to effectively exploit the two-dimensional structure of the input data. To avoid the existence of billions of parameters if all layers are fully connected, the concept of shared weight in convolutional layers has been introduced, whereby the same filter is used for each patch in the layer; this reduces the required memory capacity and improves performance. A CNN can be trained using a back-propagation algorithm [23]. Compared with other deep learning structures, a CNN gives better results in applications such as image and voice recognition.

In this study, a CNN was used to design a network for facial beauty prediction. We randomly selected 400 images from our SCUT-FBP dataset for training, and the remaining 100 images were used for testing. The network outputs a score for each test face. The correlation between the preset score and the predicted score was used to evaluate the network.

We designed a convolutional neural network for facial beauty prediction; this network contained six convolution layers, each of which was followed by a max-pooling layer. The numbers of feature maps applied to the six convolution layers were 50, 100, 150, 200, 250, and 300; the sizes of the corresponding filters were are 5×5, 5×5, 4×4, 4×4, 4×4, and 2×2. Such a combination was found to give better results than networks with a greater number of feature maps or smaller filters. There were two fully connected layers at the top of the network: the first one had 500 neurons, whereas the second one had only one neuron because we wanted it to output the predicted score of the input image. To enhance the network, we used some tricks such as dropout. Finally, the Euclidean loss was selected as the loss function. The architecture of the network is shown in Figure 8.

We conducted five experiments using five types of randomly selected training and test sets, and we calculated the correlation coefficient for each of them. In addition, we calculated the average correlation coefficient. The results are listed in Table VI.

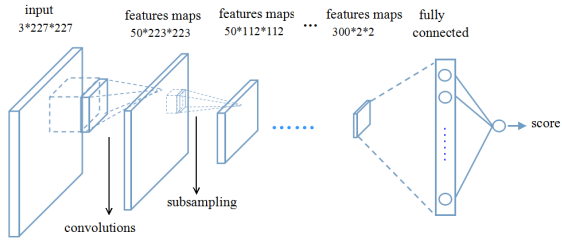

Figure 8. Network architecture of our CNN for facial beauty prediction

TABLE VI.    CORRELATION COEFFICIENTS IN SINGLE NETWORK

| *Exp.* | 1 | 2 | 3 | 4 | 5 | Average |
|---|---|---|---|---|---|---|
| PC | 0.8509 | 0.8050 | 0.8112 | 0.7817 | 0.8446 | 0.8187 |

In the case of a single network, we obtained an average correlation coefficient of 0.8187, indicating a good correlation between the preset scores and the predicted scores obtained by CNN. This indicates that the CNN-based deep learning approach shows good performance for facial beauty prediction.

V. CONCLUSION

We developed a dataset of faces with attractiveness ratings, namely, the SCUT-FBP dataset. This dataset contains face portraits of 500 Asian female subjects with attractiveness ratings, and it is publicly available at http://www.hcii-lab.net/data/SCUT-FBP/. We analyzed and verified the facial attractiveness ratings from many aspects, thereby confirming the reliability of the dataset. In addition, we presented a benchmark evaluation based on traditional machine learning and deep learning approaches. The best Pearson correlation (0.8187) was achieved by the CNN model. The SCUT-FBP dataset can be used to investigate different aspects of facial beauty analysis problems and thus promote further development in this field.

ACKNOWLEDGEMENT

This research was supported in part by the National Natural Science Foundation of China (Grant No.: 61472144), the National Science and Technology Support Plan (Grant No.: 2013BAH65F01-2013BAH65F04), and the Research Fund for the Doctoral Program of Higher Education of China (Grant No.: 201120172110023).

REFERENCES

[1]  R. Thornhill and S. W. Gangestad, "Facial attractiveness," Trends Cognit.Sci., vol. 3, no. 12, pp. 452–460, 1999.
[2]  Y. Eisenthal, G Dror, and E. Ruppin, "Facial attractiveness: Beauty and the machine," Neural Computation, vol. 18, pp. 119-142, 2006.
[3]  H. Mao, L. Jin, and M. Du, "Automatic classification of Chinese female facial beauty using Support Vector Machine," in Proc. IEEE SMC, pp. 4842-4846, 2009.
[4]  F. Chen and D. Zhang, "A benchmark for geometric facial beauty study," Medical biometrics. Springer Berlin Heidelberg, pp. 21-32, 2010.
[5]  Etcoff, and L. Nancy, "Beauty and the beholder," Nature, vol.368, no. 6468, pp. 186–187, 1994.
[6]  Langlois, H. Judith, and L.A. Roggman, "Attractive Faces are only Average," Psychological Science, vol. 1, pp. 115–121, 1990.
[7]  H. Mao, "Feature Analysis and Machine Learning of Facial Beauty Attractiveness," Ph.D. Dissertation of South China Univ. of Tech., 2011.
[8]  R. White, A. Eden, and M. Maire, "Automatic Prediction of Human Attractiveness," UC Berkley CS280A Project, vol. 1, pp. 2, 2004.
[9]  D. Zhang, Q, Zhao, and F. Chen, "Quantitative analysis of human facial beauty using geometric features," Pattern Recognition, vol. 44, no. 4, pp. 940-950, 2011.
[10] B.H. Cohen, "Explaining psychological statistics," John Wiley & Sons, 2008.
[11] J. Gan, Li L, Y. Zhai, and Y. Liu, "Deep self-taught learning for facial beauty prediction," Neurocomputing, vol. 144, pp. 295-303, 2014.
[12] J. Fan J, K.P. Chau, K P, X. Wan, L. Zhai and E. Lau, "Prediction of facial attractiveness from facial proportions,".Pattern Recognition, vol. 45, no. 6, pp. 2326-2334, 2012.
[13] H. Yan, "Cost-sensitive ordinal regression for fully automatic facial beauty assessment," Neurocomputing, vol. 129, pp. 334-342, 2014.
[14] M. Redi, N. Rasiwasia, G. Aggarwal, A. Jaimes, "The Beauty of Capturing Faces: Rating the Quality of Digital Portraits," arXiv preprint arXiv:1501.07304, 2015.
[15] N. Murray, L. Marchesotti L, and F. Perronnin, "AVA: A large-scale database for aesthetic visual analysis," in Proc. CVPR, 2012.
[16] B.C. Davis and S. Lazebnik, "Analysis of human attractiveness using manifold kernel regression," in Proc. ICIP 2008.
[17] H. Gunes and M. Piccardi, "Assessing facial beauty through proportion analysis by image processing and supervised learning," International journal of human-computer studies, vol. 64, no. 12, pp. 1184-1199, 2006.
[18] K. Schmid, D. Marx, and A Samal, "Computation of a face attractiveness index based on neoclassical canons, symmetry, and golden ratios," Pattern Recognition, vol. 41, no. 8, pp. 2710-2717, 2008.
[19] J. Whitehill and J.R. Movellan, "Personalized facial attractiveness prediction," Automatic Face & Gesture Recognition, pp. 1-7, 2008.
[20] B.S. Atiyeh and S.N. Hayek, "Numeric expression of aesthetics and beauty," Aesthetic plastic surgery, vol. 32, no. 2, pp. 209-216, 2008.
[21] A.H Iliffe, "A study of preferences in feminine beauty," British Journal of Psychology, vol. 54, no. 3, pp. 267-273, 1960.
[22] G. Rhodes, "The evolutionary psychology of facial beauty," Annu. Rev. Psychol., vol. 57, pp. 199-226, 2006.
[23] D.E. Rumelhart, G.E. Hinton, and R.J. Williams, "Learning representations by back-propagating errors," Cognitive modeling, 1988.
[24] "Meitu," url:http:// xiuxiu.web.meitu.com.
[25] "ArcSoft, PorTrait+," url:http://www.arcsoft.com/ portraitplus/
[26] "ShutterStock" url: www.shutterstock.com/.
[27] "DataTang" url: http://datatang.com/.
[28] "Pixel Solutions" url: http://www.pixelall.com/.
[29] J. Whitehill, G. Littlewort, I. Fasel I, M. Bartlett, and J. Movellan, "Developing a practical smile detector," in Proc. IEEE Int. Conf. Automatic Face and Gesture Recognition, 2008.
[30] K. Pearson, "Notes on the history of correlation," Biometrika, pp. 25-45, 1920.
[31] R.Y. Brown Beatty, "An Analysis of Racial Identity Attitudes and the Perception of Racial Climate on Job Satisfaction of African American Faculty at Historically White Institutions," Ohio University, 2012.
[32] L. Liang, L. Jin and X. Li, "Facial skin beautification using adaptive region-aware mask," IEEE Trans. on Cybernetics, vol.44, no.12, pp.2600-2612, 2014.